\documentclass[letterpaper, 10 pt, conference]{ieeeconf}  %
\IEEEoverridecommandlockouts                              %
\usepackage[english]{babel}
\usepackage[T1]{fontenc}

\usepackage[nolist]{acronym}
\usepackage[cmex10]{amsmath}
\usepackage{amssymb}
\usepackage{bm}
\usepackage{booktabs}
\usepackage{color}
\usepackage{cite}
\usepackage{float}
\usepackage{graphicx}
\usepackage{hyperref}
\usepackage{import}
\usepackage{mathtools}
\usepackage{multirow}
\usepackage{outline}
\usepackage{paralist}
\usepackage{siunitx}
\usepackage{stmaryrd}
\usepackage{systeme}
\usepackage[font={small}]{caption}
\usepackage{subcaption}
\usepackage{tikz}
\usepackage{url}
\usetikzlibrary{external}

\usepackage{pgfplots, pgfplotstable}

\usetikzlibrary{positioning}
\usetikzlibrary{calc}

\definecolor{my_blue}{rgb}{0, 0.4470, 0.7410}
\definecolor{my_yellow}{rgb}{0.9290, 0.6940, 0.1250}
\definecolor{my_purple}{rgb}{0.4940, 0.1840, 0.5560}
\definecolor{my_green}{rgb}{0.4660, 0.6740, 0.1880}
\definecolor{my_red}{rgb}{0.6350, 0.0780, 0.1840}
\definecolor{my_black}{rgb}{0.25, 0.25, 0.25}

\pgfplotsset{compat=newest}

\pgfplotsset{
    legend style={font=\footnotesize},
    label style={font=\footnotesize, inner sep=0pt},
    legend cell align={left},
    tick label style={font=\footnotesize},
    mesh line legend/.style={legend image code/.code=\meshlinelegend#1}
    }

\makeatletter
\long\def\meshlinelegend#1{%
    \scope[%
        #1,
        /pgfplots/mesh/rows=1,
        /pgfplots/mesh/cols=4,
        /pgfplots/mesh/num points=,
        /tikz/x={(0.44237cm,0cm)}, 
        /tikz/y={(0cm,0.23932cm)},
        /tikz/z={(0.0cm,0cm)},
        scale=0.4,
    ]
    \let\pgfplots@metamax=\pgfutil@empty
    \pgfplots@curplot@threedimtrue

    \pgfplotsplothandlermesh
    \pgfplotstreamstart

    \def\simplecoordinate(##1,##2,##3){%
        \pgfmathparse{1000*(##3)}%
        \pgfmathfloatparsenumber\pgfmathresult
        \let\pgfplots@current@point@meta=\pgfmathresult
        \pgfplotstreampoint{\pgfqpointxyz@orig{##1}{##2}{##3}}%
    }%

    \simplecoordinate(0,0,0)
    \simplecoordinate(0.25,0,0.125)
    \simplecoordinate(0.5,0,0.25)
    \simplecoordinate(0.75,0,0.375)
    \simplecoordinate(1,0,0.5)
    \simplecoordinate(1.25,0,0.625)
    \simplecoordinate(1.5,0,0.75)
    \simplecoordinate(1.75,0,0.875)
    \simplecoordinate(2,0,1)

    \pgfplotstreamend
    \pgfusepath{stroke}
    \endscope
}%
\makeatother

\newcommand{\plotuturnball}[1]{
\begin{tikzpicture}
\begin{axis}[
    height=1*#1,
    mark size=2.5,
    enlargelimits=0.05,
    x dir=reverse,
    xlabel = {Position y [\SI{}{\m}]}, 
    ylabel = {Position x [\SI{}{\m}]},
    legend style={at={(0.7,0.3)}, font=\scriptsize, inner sep=1pt},
    colormap/viridis,
    inner sep=1pt,
    colorbar right,
    colorbar style={
        inner sep=1pt,
        width=5pt,
        title = Time [s],
        title style = {at={(0.5,-0.12)}, anchor=north, font=\footnotesize}
    },
    axis equal image,
    point meta rel=per plot,
    point meta=\thisrow{time}]
    
    \def\offsetx{-5.0}
    \def\offsety{-8.0}
    \def\cosyaw{cos(\thisrow{/odom/pose/pose/orientation/yaw_deg})}
    \def\sinyaw{sin(\thisrow{/odom/pose/pose/orientation/yaw_deg})}
    \def\velx{\thisrow{/odom/twist/twist/linear/x}}
    \def\vely{\thisrow{/odom/twist/twist/linear/y}}

    \addplot[scatter, only marks, opacity=0.6,
        x filter/.code={ %
            \pgfmathparse{mod(\thisrow{/ball_global/header/seq},30) < 1 ? x : nan}
        }] 
        table[col sep=comma, x expr=\thisrow{/ball_global/point/y_ma}+\offsetx, y expr=\thisrow{/ball_global/point/x_ma}+\offsety]
        {data/u_turn_ball.csv};
        \addlegendentry{Ball}

    \addplot[mesh, mesh line legend, line width=1pt]
        table[col sep=comma, x expr=\thisrow{/odom/pose/pose/position/y}+\offsetx, y expr=\thisrow{/odom/pose/pose/position/x}+\offsety] 
        {data/u_turn_odom_50_downsampled.csv};
        \addlegendentry{Robot}

    \addplot[quiver={
                v=\cosyaw*\velx+\sinyaw*\vely,
                u=\sinyaw*\velx-\cosyaw*\vely,
                scale arrows=1.0, colored}, 
        -stealth,
        line width=0.5,
        x filter/.code={ %
            \pgfmathparse{mod(\thisrow{/odom/header/seq},4) < 1 ? x : nan}
        }]
        table[col sep=comma, x expr=\thisrow{/odom/pose/pose/position/y}+\offsetx, y expr=\thisrow{/odom/pose/pose/position/x}+\offsety] 
        {data/u_turn_odom_50_downsampled.csv};

    \draw[->, thick, font=\scriptsize] (0.1,2) -- (0.5,2) node[anchor=east, align=center] {1st Turn\\Cmd};
    \draw[->, thick, font=\scriptsize] (2,4.2) -- (2,3.7) node[anchor=north, align=center] {2nd Turn\\Cmd};
\end{axis}
\end{tikzpicture}
}

\newcommand{\plotuturndirection}[2]{
    \begin{tikzpicture}
        \begin{axis}[
            width=#1,
            height=#2,
            ymin = -40,
            enlargelimits=false,
            xlabel = {Time [\SI{}{\second}]}, 
            ylabel = {Velocity Direction [\SI{}{\degree}]},
            ytick={0,90,180},
            align=center,
            inner sep=1pt,
            legend pos = south east,
            legend style={font=\scriptsize, inner sep=1pt},
            ]

            \addplot[my_blue, only marks, mark size=1.0, opacity=0.4,
            y filter/.code={
                \pgfmathparse{abs(\thisrow{/ball_global/point/x_vel_ma}) + abs(\thisrow{/ball_global/point/y_vel_ma})}
                \pgfmathparse{\pgfmathresult < 0.15 ? nan : \thisrow{/ball_global/point/direction}}
                \pgfmathparse{\pgfmathresult < -2.8 ? \pgfmathresult+6.28 : \pgfmathresult},
                \pgfmathparse{\pgfmathresult/3.1415*180}
                }] 
                table[col sep=comma, x expr=\thisrow{time}] 
                {data/u_turn_ball.csv};
    
            \addplot[my_yellow, line width=2pt] 
                table[col sep=comma, x expr=\thisrow{time}, y expr=\thisrow{/odom/pose/pose/orientation/yaw_deg}] 
                {data/u_turn_odom_50_downsampled.csv};

            \addplot[my_purple, line width=2pt, dashed]
                coordinates {
                    (0.,0.) (9.6,0.) (9.6,90) (18.1,90) (18.1,180) (31.5,180)
                };
            \legend{Ball, Robot, Command}

            \draw[->, thick, font=\footnotesize] (13,135) node[anchor=east] {Detection\\Error} -- (15.8,127);

        \end{axis}
    \end{tikzpicture}
}

\newcommand{\plotpushbox}{
\def\x{\thisrow{/uwb_odom/pose/pose/position/x}}
\def\y{\thisrow{/uwb_odom/pose/pose/position/y}}
\def\yaw{\thisrow{/uwb_odom/pose/pose/orientation/yaw_deg}}
\def\cosyaw{cos(\yaw)}
\def\sinyaw{sin(\yaw)}
\begin{tikzpicture}
\begin{axis}[
    name=plot,
    height=0.28\linewidth,
    mark size=4.0,
    enlargelimits=0.05,
    xlabel = {Position x [\SI{}{\m}]}, 
    ylabel = {Position y [\SI{}{\m}]},
    label style={inner sep=-2pt},
    xmin=0.8,
    xmax=3.5,
    ymin=1.3,
    ymax=3.5,
    colormap/viridis,
    axis equal image,
    point meta=\thisrow{time}]
    
    \def\offsetx{-0.0}
    \def\offsety{-0.0}

    \addplot[quiver={
                u=\cosyaw+\sinyaw,
                v=\sinyaw-\cosyaw,
                scale arrows=0.1, colored}, 
        -stealth,
        line width=0.5,
        x filter/.code={ %
            \pgfmathparse{mod(\thisrow{/uwb_odom/header/seq},8) < 1 ? x : nan}
        }]
        table[col sep=comma, x expr=\x+\offsetx, y expr=\y+\offsety] 
        {data/push_box_baseline_downsampled.csv};
    
    \addplot[mesh, line width=1pt]
        table[col sep=comma, x expr=\x+\offsetx, y expr=\y+\offsety] 
        {data/push_box_baseline_downsampled.csv};

    \addplot[scatter, mark=square*, only marks, opacity=0.6] 
    table[col sep=comma, row sep=crcr, x expr=\thisrow{x}+\offsetx, y expr=\thisrow{x}+\offsety]{
        time,x,y\\
        0,1.842,1.787\\
        5,1.738,1.795\\
        15,1.87,2.04\\
        20,1.99,2.02\\
        23,2.02,2.12\\
        25.,2.30,2.32\\
        38.9,2.497,2.348\\
        43.9,2.7,2.6\\
    };
    
    \addplot[my_red, fill=my_red, opacity=0.6,domain=0:356.4,samples=101,smooth cycle,point meta=none] ({3.24+0.6*cos(x)},{2.5+0.6*sin(x)});
    \node[my_red] at (axis cs: 3.24,2.5) {Goal};
    \node[draw,my_yellow,line width=1pt,name=highlight,minimum width=1.5cm, minimum height=0.4cm] at (axis cs:1.5, 1.8) {};
\end{axis}

\node[xshift=-0.04\linewidth,left=0pt of plot, anchor=east, inner sep=0pt] (real) {\includegraphics[width=0.18\linewidth,trim={150 100 550 0},clip]{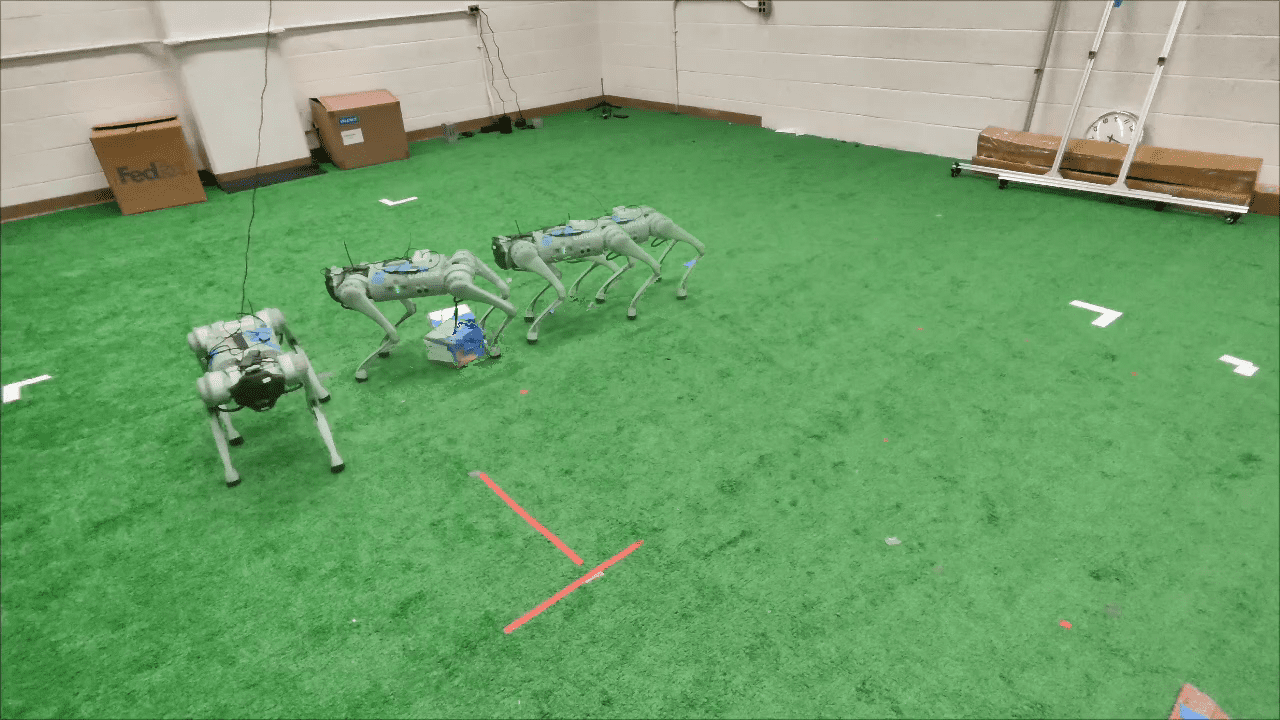}};
\draw[my_yellow,line width=1pt] (real.south west) rectangle (real.north east);
\draw[my_yellow, dashed, line width=1pt] (highlight.north west) -- (real.north east);
\draw[my_yellow, dashed, line width=1pt] (highlight.south west) -- (real.south east);
\node[black, anchor=north east] {(a)};

\end{tikzpicture}
\hspace{10pt}
\begin{tikzpicture}
\begin{axis}[
    name=plot,
    height=0.28\linewidth,
    mark size=4,
    enlargelimits=0.05,
    xlabel = {Position x [\SI{}{\m}]}, 
    ylabel = {Position y [\SI{}{\m}]},
    label style={inner sep=-2pt},
    xmin=0.8,
    xmax=3.5,
    ymin=1.3,
    ymax=3.5,
    colormap/viridis,
    colorbar right,
    colorbar style={
        inner sep=1pt,
        width=5pt,
        title = Time [s],
        title style = {at={(0.5,-0.1)}, anchor=north, font=\footnotesize}
    },
    axis equal image,
    legend pos=north west,
    point meta=\thisrow{time}]

    \def\offsetx{-0.0}
    \def\offsety{-0.0}

    \addplot[mesh, mesh line legend, line width=1pt]
        table[col sep=comma, x expr=\x+\offsetx, y expr=\y+\offsety] 
        {data/push_box_finetune_downsampled.csv};
        \addlegendentry{Robot}

    \addplot[scatter, mark=square*, only marks, opacity=0.6] 
    table[col sep=comma, row sep=crcr, x expr=\thisrow{x}+\offsetx, y expr=\thisrow{x}+\offsety]{
        time,x,y\\
        0,1.771,1.755\\
        7.55,2.039,1.655\\
        9.50,2.10,1.58\\
        12.56,2.456,2.265\\
        23.83,2.6509,2.7029\\
        44.383,0,0\\ %
    };
    \addlegendentry{Load Position}

    \addplot[quiver={
                u=\cosyaw+\sinyaw,
                v=\sinyaw-\cosyaw,
                scale arrows=0.1, colored}, 
        -stealth,
        line width=0.5,
        x filter/.code={ %
            \pgfmathparse{mod(\thisrow{/uwb_odom/header/seq},8) < 1 ? x : nan}
        }]
        table[col sep=comma, x expr=\x+\offsetx, y expr=\y+\offsety] 
        {data/push_box_finetune_downsampled.csv};
    
    \addplot[my_red, fill=my_red, opacity=0.6, domain=0:356.4,samples=101,smooth cycle,point meta=none] ({3.24+0.6*cos(x)},{2.5+0.6*sin(x)});
    \node[my_red] at (axis cs: 3.24,2.5) {Goal};
    \node[draw,my_yellow,line width=1pt,name=highlight,minimum width=0.55cm, minimum height=0.75cm] at (axis cs:1.6, 2.0) {};
\end{axis}

\node[xshift=-0.04\linewidth,left=0pt of plot, anchor=east, inner sep=0pt] (real) {\includegraphics[width=0.18\linewidth,trim={0 0 700 100},clip]{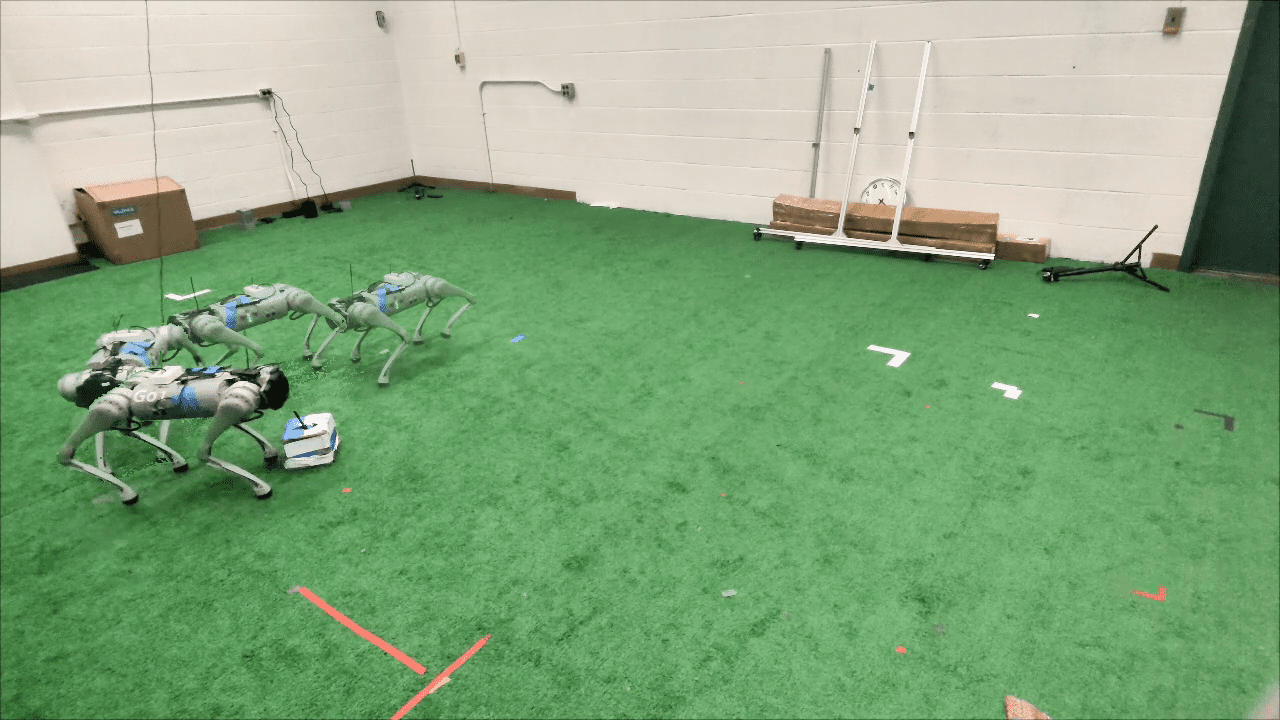}};
\draw[my_yellow,line width=1pt] (real.south west) rectangle (real.north east);
\draw[my_yellow, dashed, line width=1pt] (highlight.north west) -- (real.north east);
\draw[my_yellow, dashed, line width=1pt] (highlight.south west) -- (real.south east);
\node[black, anchor=north east] {(b)};

\end{tikzpicture}
}

\newcommand{\figuturn}{

\begin{figure*}[ht]
    \centering
    \subfloat[Recorded ball, robot position, and robot velocity trajectory]{
        \plotuturnball{0.2\textheight}
    }
    \hfill
    \subfloat[Recorded velocity direction data.]{\plotuturndirection{0.42\linewidth}{0.2\textheight}}
    \hfill
    \subfloat[Snapshots of quadruped dribbling the ball.]{\includegraphics[height=0.155\textheight,trim={250 0 0 50},clip]{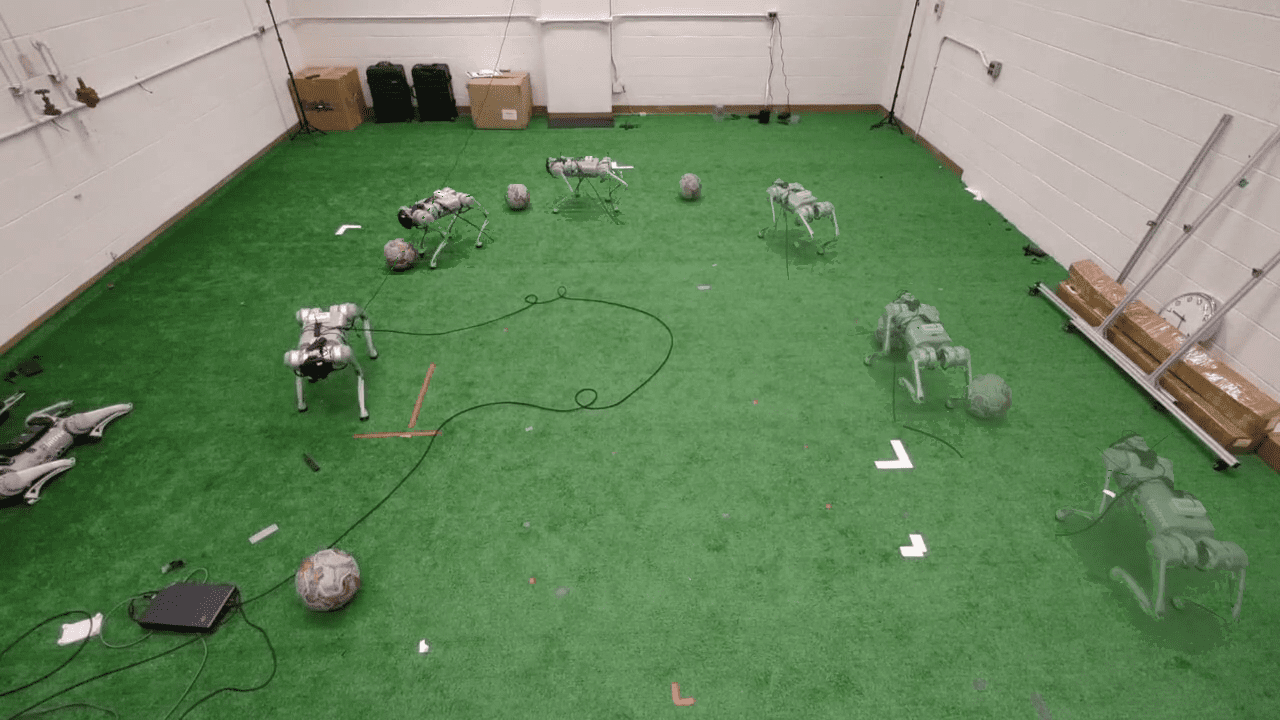}}
    \caption{Visualization of real-world ball dribbling experiments, using the proposed HILMA-Res framework. The quadruped can perform a sharp U-Turn within a narrow space of less than 3.3 meters in width. This demonstrates that the HILMA-Res can enable agile loco-manipulation maneuvers that can be directly transferred from simulation to the real world.}
    \label{fig:uturn}
    \vspace{-2pt}
    \end{figure*}
}

\newcommand{\figpushbox}{
\begin{figure*}[t]
    \plotpushbox
    \caption{Recorded data during \emph{NavLoad} experiments, along with snapshots capturing the robot's behavior during its first interaction with the load. (a) Zero-shot transfer of the base policy trained in simulation, the robot tends to have a large detour to move the load to the target due to a large sim-to-real gap (such as friction, sensor noise, etc). (b) Training with real-world data by RLPD, the robot first adjusts its pose and then pushes the load along the direction of the goal, with shorter path and operation time. This showcases the advantages of HILMA-Res in the fine-grained manipulation task that requires efficient training from real-world data.}
    \label{fig:pushbox}
\end{figure*}

}

\newcommand{\figstepoverstone}{
\def\mainwidth{0.95\linewidth}
\def\zoomheight{0.078\textheight}
\def\smallgap{5pt}
\begin{figure*}[t]
    \centering
    \begin{tikzpicture}[line width=1.5pt,inner sep=0pt]
        \node (main1) at (0,0) {\includegraphics[width=\mainwidth]{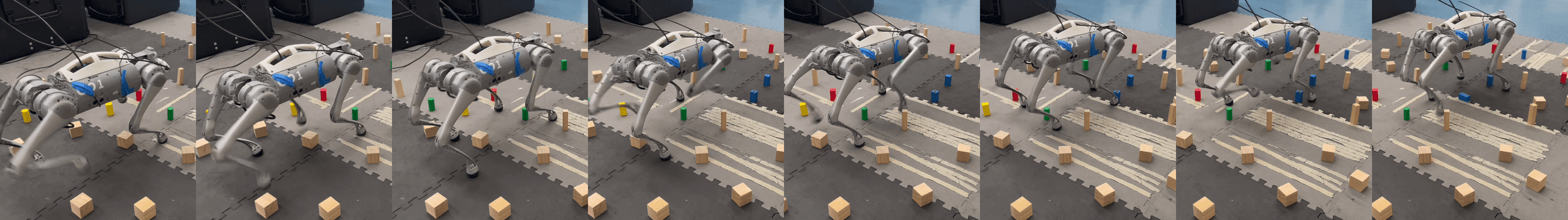}};
        \foreach \i in {1,...,7}
        \draw [white] ($(main1.south west)!\i*0.125!(main1.south east)$) -- ($(main1.north west)!\i*0.125!(main1.north east)$);

        \node[xshift=-0.23\linewidth,below=\smallgap of main1] (zoom1) {\includegraphics[height=\zoomheight,trim={0 150 0 50},clip]{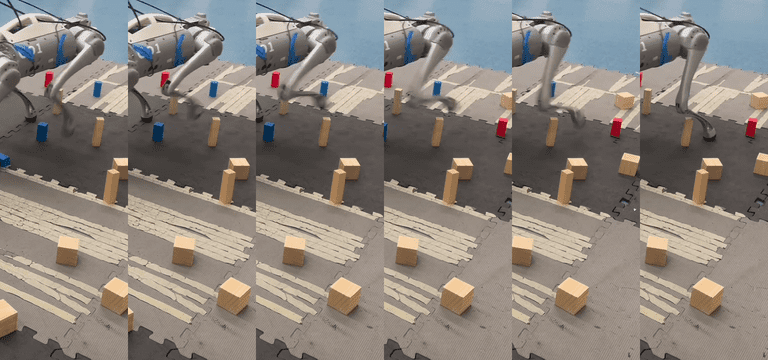}};
        \foreach \i in {1,...,5}
        \draw [my_green] ($(zoom1.south west)!\i*0.166!(zoom1.south east)$) -- ($(zoom1.north west)!\i*0.166!(zoom1.north east)$);

        \draw[my_green] ($(main1.south east)!0.1!(main1.south west)$) rectangle (main1.north east);
        \draw[my_green] (zoom1.south west) rectangle (zoom1.north east);
        \draw[my_green, dashed] (zoom1.north west) -- ($(main1.south east)!0.1!(main1.south west)$);
        \draw[my_green, dashed] (zoom1.north east) -- (main1.south east);
        \node[align=center, yshift=-0.4*\zoomheight, white, rectangle, fill=my_green, inner sep=2pt] at (zoom1.center) {The front swing leg avoiding stone};

        \node[below=\zoomheight+2*\smallgap of main1] (main2) {\includegraphics[width=\mainwidth,trim={0 0 0 0},clip]{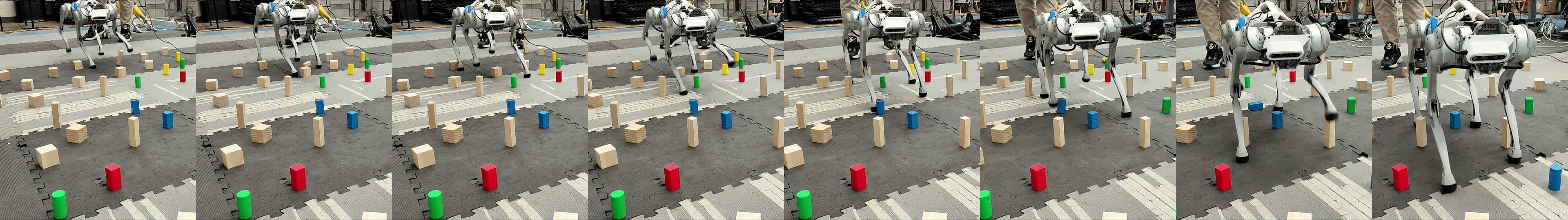}};
        \foreach \i in {1,...,7}
        \draw [white] ($(main2.south west)!\i*0.125!(main2.south east)$) -- ($(main2.north west)!\i*0.125!(main2.north east)$);

        \node[right=5pt of zoom1] (zoom2) {\includegraphics[height=\zoomheight,trim={0 50 0 150},clip]{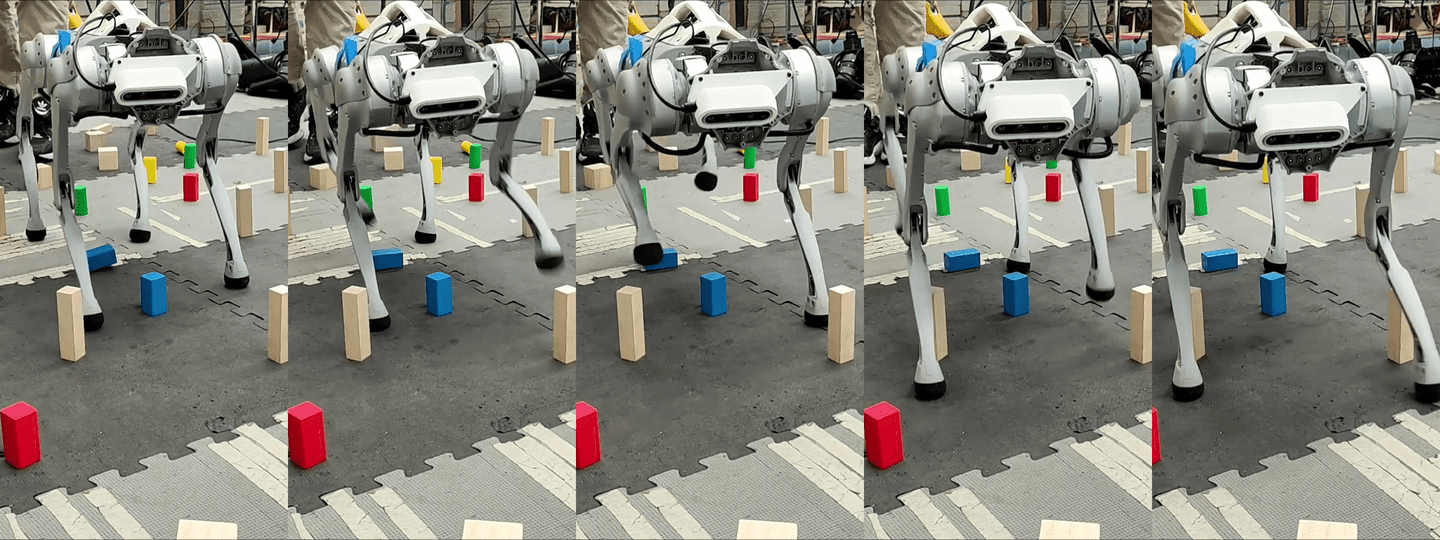}};
        \foreach \i in {1,...,4}
        \draw [my_purple] ($(zoom2.south west)!\i*0.2!(zoom2.south east)$) -- ($(zoom2.north west)!\i*0.2!(zoom2.north east)$);

        \draw[my_purple] ($(main2.south east)!0.14!(main2.south west)$) rectangle (main2.north east);
        \draw[my_purple] (zoom2.south west) rectangle (zoom2.north east);
        \draw[my_purple, dashed] (zoom2.south west) -- ($(main2.north east)!0.14!(main2.north west)$);
        \draw[my_purple, dashed] (zoom2.south east) -- (main2.north east);
        \node[align=center, yshift=-0.4*\zoomheight, white, rectangle, fill=my_purple, inner sep=2pt] at (zoom2.center) {The front swing leg avoiding stone};

    \end{tikzpicture}
    \caption{Snapshots of two real-world experiments on \emph{StepOStone}. The robot can walk through the randomly positioned small cluttered stones, employing a dynamic trotting gait. The zoom-in images emphasize the robot's ability to sidestep obstacles by adjusting its swing legs. Note that some of the fall-down stones are caused by the impact variation when the legs make contact with the ground.}
    \label{fig:stepostone}
    \vspace{-3pt}
\end{figure*}
}

\newcommand{\figcover}{
\begin{figure}
    \def\width{0.31\linewidth}
    \def\smallgap{0pt}
    \centering
    \begin{tikzpicture}[inner sep=0pt, font=\footnotesize]
        \node (dribbling) {\includegraphics[width=\width]{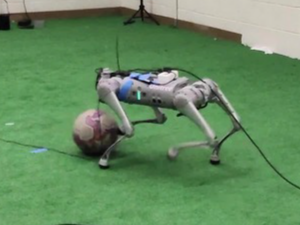}};
        \node[right=\smallgap of dribbling] (stepoverstone) {\includegraphics[width=\width]{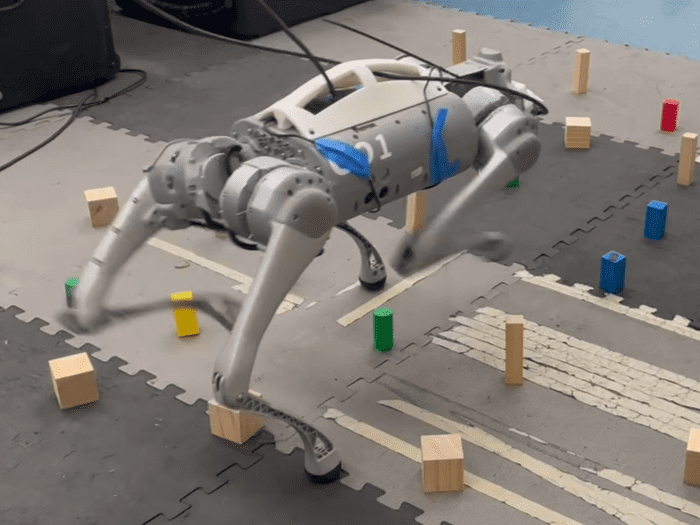}};
        \node[right=\smallgap of stepoverstone] (pushbox) {\includegraphics[width=\width]{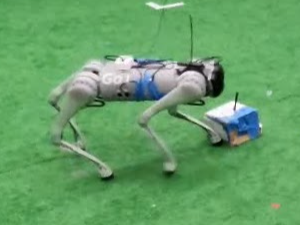}};
        \node[white, anchor=south west, inner sep=2pt] at (dribbling.south west) {Ball Dribbling};
        \node[white, anchor=south west, inner sep=2pt] at (stepoverstone.south west) {Stepping over Stones};
        \node[white, anchor=south west, inner sep=2pt] at (pushbox.south west) {Navigating Load};
    \end{tikzpicture}
    \caption{The proposed HILMA-Res framework enables a quadrupedal robot to perform different loco-manipulation tasks in the real world. These include dribbling a ball in a desired direction, stepping over small blocks scattered on the ground, and navigating a load to the desired goal through real-world learning. We highlight the versatility of the HILMA-Res framework in various loco-manipulation tasks with different observation spaces and learning algorithms. }\label{fig:intro}
    \vspace{-0.2cm}
\end{figure}
}

\title{\LARGE \bf HiLMa-Res: A General Hierarchical Framework via Residual RL for Combining Quadrupedal Locomotion and Manipulation}
\author{Xiaoyu Huang$^{1}$, Qiayuan Liao$^{1}$, Yiming Ni$^{1}$, Zhongyu Li$^1$, Laura Smith$^1$, \\ Sergey Levine$^1$, Xue Bin Peng$^2$, Koushil Sreenath$^1$
\thanks{$^1$ University of California, Berkeley, USA}
\thanks{$^2$ Simon Fraser University, Canada}
\thanks{Email: haytham.huang@berkeley.edu}}

\begin{document}
\maketitle

\begin{abstract}
This work presents HiLMa-Res, a hierarchical framework leveraging reinforcement learning to tackle manipulation tasks while performing continuous locomotion using quadrupedal robots. Unlike most previous efforts that focus on solving a specific task, HiLMa-Res is designed to be general for various loco-manipulation tasks that require quadrupedal robots to maintain sustained mobility. The novel design of this framework tackles the challenges of integrating continuous locomotion control and manipulation using legs. It develops an operational space locomotion controller that can track arbitrary robot end-effector (toe) trajectories while walking at different velocities. This controller is designed to be general to different downstream tasks, and therefore, can be utilized in high-level manipulation planning policy to address specific tasks. To demonstrate the versatility of this framework, we utilize HiLMa-Res to tackle several challenging loco-manipulation tasks using a quadrupedal robot in the real world. These tasks span from leveraging state-based policy to vision-based policy, from training purely from the simulation data to learning from real-world data. In these tasks, HiLMa-Res shows better performance than other methods. 

\end{abstract}

\section{Introduction}\label{sec:intro}
Using legs as manipulators to perform non-prehensile manipulation tasks is a natural behavior commonly observed among humans. 
Beyond just locomotion, people are capable of using their legs for a variety of actions, such as dribbling a soccer ball while walking or running, or moving an object by kicking it rather than lifting.
Enabling legged robots, like quadrupeds, to perform such loco-manipulation tasks with their legs while moving could significantly enhance their versatility and applicability.
However, this is a hard problem. Using the quadrupedal robot as an example, the robot needs to consider its stability while walking and also has to adjust its legs and use whole-body maneuvers to accomplish fine-grained manipulation tasks. 
The robot has to update its leg movement strategy not only for locomotion control but also for manipulation planning in real-time.
Developing a general solution for different loco-manipulation tasks further complicates this due to the distinct objectives and environments associated with different manipulation tasks.  
Many attempts have been made to address the challenge of loco-manipulation, which involves the coupled problem of locomotion control and manipulation planning. Due to its difficulty, many efforts have opted to simplify the scenario. 
This simplification includes focusing on a single task, such as just ball dribbling~\cite{bohez2022imitate,ji2023dribblebot}, or developing separate controllers like~\cite{cheng2023legs,ji2022hierarchical,huang2023creating}: one for manipulation using the robot's legs and another for continuous locomotion skills.
It is still an open question to develop a general framework that can solve different manipulation tasks using the robot's legs while performing continuous locomotion. 
In this work, we propose a hierarchical framework, named HiLMa-Res, to divide the complex loco-manipulation task and conquer the control and planning sub-problems individually by reinforcement learning (RL). It can be applied to different loco-manipulation tasks in the real world, as shown in Fig.~\ref{fig:intro}.

\figcover

\subsection{Contributions}
The main contribution of this work is the introduction of HiLMa-Res. (1) It is a general framework enabling quadrupedal robots to perform various manipulation tasks while maintaining sustained mobility through hierarchical reinforcement learning (RL). (2) The novel design of HiLMa-Res addresses the challenges of integrating manipulation skills with continuous locomotion. This includes implementing a single task-independent operational space locomotion controller, such as a walking controller, to track arbitrary end-effector (toe) trajectories, and a high-level task-specific manipulator planner for determining a residual trajectory for the end-effector.
(3) We demonstrate that this design offers a versatile solution for different loco-manipulation tasks, requiring minimal effort for retraining on individual tasks. 
This framework supports different observation spaces and RL algorithms. 
(4) Our experiments showcase the ability of HiLMa-Res by realizing various loco-manipulation tasks in the real world, including ball dribbling (training from simulation data), stepping over stones (using visuomotor skills), and load navigation (training from real-world data). Additionally, we show that HiLMa-Res outperforms other state-of-the-art RL methods in the challenging loco-manipulation task.

\section{Related Work}\label{sec:related_work}
Different from the efforts on mobile manipulation with wheeled bases which does not need to consider the robot's stability~\cite{bohren2011towards, fu2024mobile}, loco-manipulation using legged robots requires different solutions. Previous work mainly focuses on two approaches: arm-equipped legged robots, and using the robot's legs as manipulators.

\emph{Mobile Manipulation with Legged Robots:}
Equipping a legged robot, like a quadruped, with an additional arm can expand its manipulation capabilities by utilizing its mobility and whole-body maneuvers. 
Previous research has explored various methods to improve the coordination between locomotion and manipulation. 
Some have proposed a residual learning framework to enhance skill coordination~\cite{yokoyama2023asc}, requiring a pre-existing locomotion controller. 
Others have focused on developing a unified whole-body policy for both manipulation and locomotion~\cite{cheng2024expressive,fu2023deep,wu2019teleoperation}, though without incorporating high-level planning for task space, and focusing solely on joint-level control. 
Model-based approaches, using predictive and whole-body control, facilitate coordinated movements by accounting for multi-rigid body dynamics but depend on predefined contact sequences and task-specific constraints~\cite{sleiman2021unified,ferrolho2023roloma}. Techniques combining trajectory optimization and contact planning have achieved versatile multi-contact loco-manipulation, albeit with the necessity for offline planning~\cite{sleiman2023versatile}. 
However, integrating an arm increases the system's complexity and necessitates additional hardware, more degrees of freedom, and the added load.

\emph{Using Legs for Both Locomotion and Manipulation:}
Attempts leveraging legs as end effectors for manipulation using legged robots could have simplicity in hardware but limit tasks to non-prehensile manipulation~\cite{gong2023legged}. 
Legs are critical for stabilizing the robot, and using them for extra manipulation tasks poses challenges in maintaining stable gaits. 
Some work avoids the need for precise manipulation by only focusing on manipulating large objects~\cite{rigo2023contact}. 
To directly use legs for manipulation tasks, some prior work used two cascaded RL-based policies for control and planning in the robot's operational space, respectively. 
This method enables a quadrupedal robot to shoot~\cite{ji2022hierarchical} or intercept~\cite{huang2023creating}.
Others opt to develop a unified policy 
However, these simplify the loco-manipulation problem by the constrained mobility, such as just standing~\cite{ji2022hierarchical} or only jumping once~\cite{huang2023creating}. 
In other work like~\cite{cheng2023legs}, a single policy is developed for individual loco-manipulation tasks that are still limited to constrained mobility, such as standing against a wall to push a button.
If there is a need for continuous locomotion, such as walking, to enlarge the robot's working space, a separate locomotion controller and rule-based policy selector are necessary in this work~\cite {ji2022hierarchical,huang2023creating,cheng2023legs}.
Specialized tasks, like dribbling soccer balls while walking, have seen some development but lack scalability~\cite{ji2023dribblebot}. 
In the humanoid robotics field, besides heuristic-based approaches like~\cite{teixeira2020humanoid}, hierarchical learning~\cite{bohez2022imitate} framework has been explored for using legs in manipulation tasks, yet this work focuses on task-specific training without a generalizable low-level controller for downstream manipulation tasks. 
This work introduces HiLMa-Res, a hierarchical framework aiming for generality across loco-manipulation tasks, addressing these limitations.

\begin{figure}[t]
  \centering
  \includegraphics[width=0.99\linewidth]{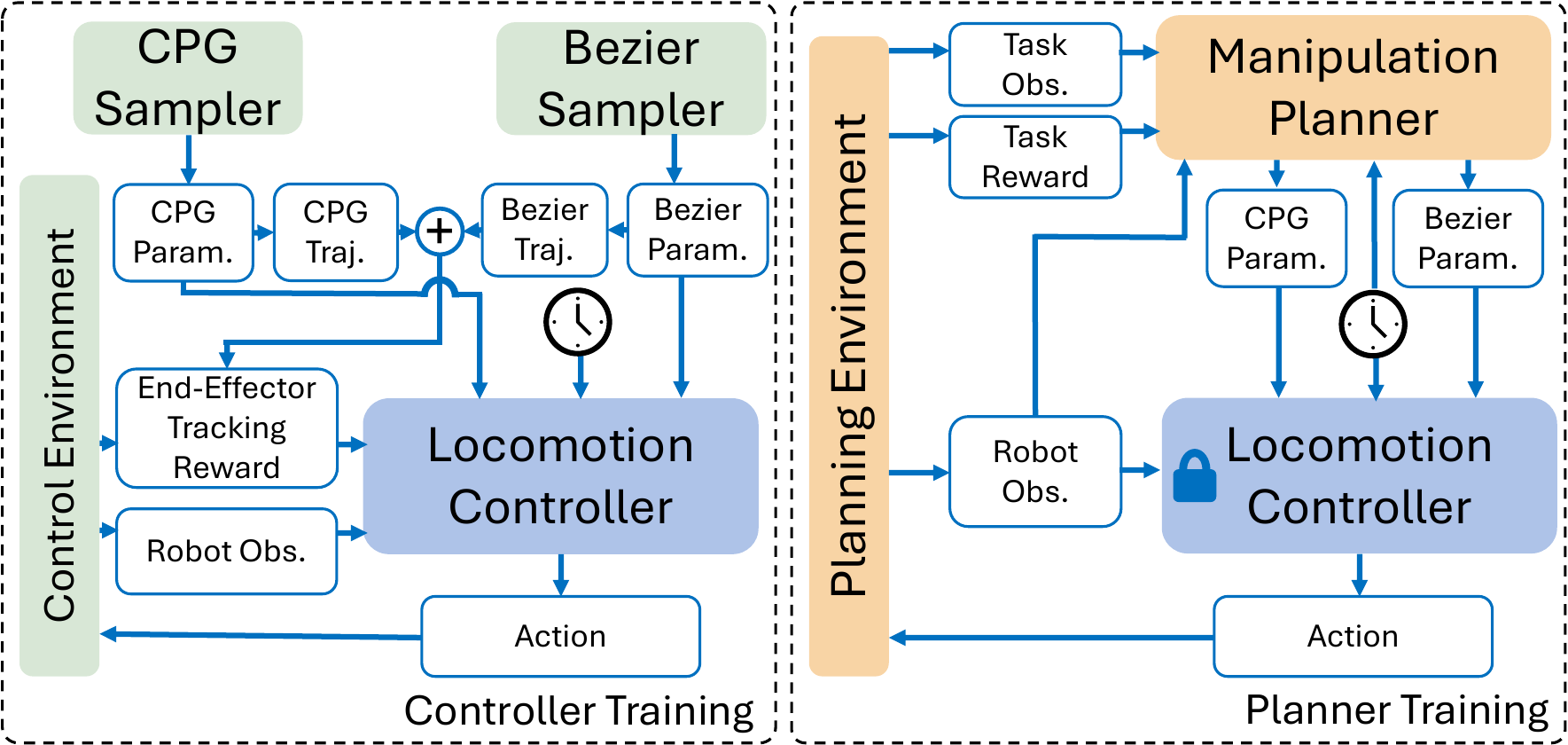}
  \caption{The HiLMa-Res framework. This hierarchical framework consists of a controller training stage where we train a task-independent locomotion controller that tracks desired end-effector trajectories. This is an addition from sampled CPG trajectories and sampled B\'ezier residual trajectories. A planner training stage is designed to reuse the locomotion controller to train a task-specific manipulation planner for downstream loco-manipulation tasks. We highlight the importance of reusing a pre-trained locomotion controller that has been evaluated in the real-world environment, which enables fast and efficient learning of the planner and prevents learning dynamically infeasible actions.}
  \label{fig:framework} 
  \vspace{-5pt}
\end{figure}

\section{The HiLMa-Res Framework and Loco-Manipulation Tasks}\label{sec:framework}
In this section, we provide an overview of the HiLMa-Res framework, as visualized in Fig.~\ref{fig:framework}. 
We frame the loco-manipulation tasks with continuous locomotion as a combination of two problems with different levels of difficulty of generalization using RL: versatile locomotion control that can be realized by a single policy and object manipulation that can be hard to generalize to different tasks.    

The HiLMa-Res contains two parts: a task-independent operational space locomotion controller and a task-specific manipulation planner, as shown in Fig.~\ref{fig:framework}. 
The locomotion controller is responsible for joint-level control of the whole body of the quadrupedal robot, leveraging the robot's proprioceptive feedback as input. 
We adopt a motion tracking method where a parameterized reference motion provided by a Central Pattern Generator (CPG)~\cite{shao2021learning} is utilized to provide gait priors for tracking various commanded velocities.  
Note that the CPG is designed to generate end-effector (toe) trajectories for the quadrupedal robot with different contact sequences, different walking velocities, and turning rates. 
This facilitates planning and control in the operational space. 
Residual trajectories, represented by B\'ezier curves, are added to the nominal end-effector trajectories of the \textit{swing legs} from CPG, while leaving the stance leg unchanged. 
The details for training this operational space locomotion controller are described in Sec.~\ref{sec:control}.
After the locomotion controller is available and evaluated in the real world, it can be kept and re-used by different task-specific planners.
The planner is designed to accomplish specified loco-manipulation tasks by specifying the residual trajectories of the end-effector and the commands for the robot moving base for the locomotion controller. 
The input of the planner includes not only the robot's proprioceptive feedback but also task-specific information, such as the object's location, or directly using depth vision as input. 
Moreover, the planner can be constructed through different methods, as described in Sec.~\ref{sec:planning}.

To demonstrate the versatility of the HiLMa-Res framework, we evaluate our method on three distinct loco-manipulation tasks, with increase in difficulty:

\textbf{Ball Dribbling (Dribble)}: In this task, a quadrupedal robot dribbles a soccer ball and controls the ball's velocity and direction while dribbling, using a vision-based ball detection with onboard cameras. 

\textbf{Stepping over Stones (StepOStone)}: This task requires a quadrupedal robot to walk through a path littered with small stones or ground obstacles without collisions, inspired by the agile movements of a cat (as seen in~\cite{obstaclechallenge}, and as tackled by a quadrupedal robot in simulation only in~\cite{kareer2023vinl}). 
Since measuring the positions of the small stone could introduce significant errors, we directly leverage the robot's onboard vision in the manipulation planner to prevent the robot's end-effectors from hitting the stones. This could allow the policy developed in simulations to be applied in the real world. 

\textbf{Navigating Load (NavLoad)}: the robot uses its front legs to push a small yet relatively heavy box towards a specified target location, requiring strategic long-term planning. The difference between simulated environments and the real world (like ground friction, size and stiffness of the load) can lead to accumulated errors over the long horizon. Therefore, this task necessitates real-world training to successfully and efficiently push the load to the target.

These tasks showcase a wide range of loco-manipulation capabilities: (1) the use of visual observation (\textbf{StepOStone}) versus state observation (\textbf{Dribble}, \textbf{NavLoad}), (2) the necessity for short-term planning (\textbf{StepOStone}, \textbf{Dribble}) versus long-term planning (\textbf{NavLoad}), and (3) tasks that demand training with real-world data (\textbf{NavLoad}) as opposed to those that can be transferred directly from simulation to the real world (\textbf{StepOStone}, \textbf{Dribble}).

\section{Task-independent Quadrupedal Locomotion Control in Operational Space}\label{sec:control}
The first component of HiLMa-Res is an RL-trained quadrupedal locomotion controller designed to be broadly applicable to a wide range of downstream loco-manipulation tasks. 
The locomotion controller is trained to enable the robot to follow a \textit{diverse} set of end-effector trajectories for the swing legs while walking, \textit{i.e.}, operating in the operational space.  
\emph{For brevity, we use “trajectories" to refer to “end-effector trajectories of the robot's swing legs".} 

\subsection{Parameterized Reference Trajectory in Operational Space}\label{subsec:ref_traj}
\begin{figure}[t]
  \centering
  \includegraphics[width=0.6\linewidth]{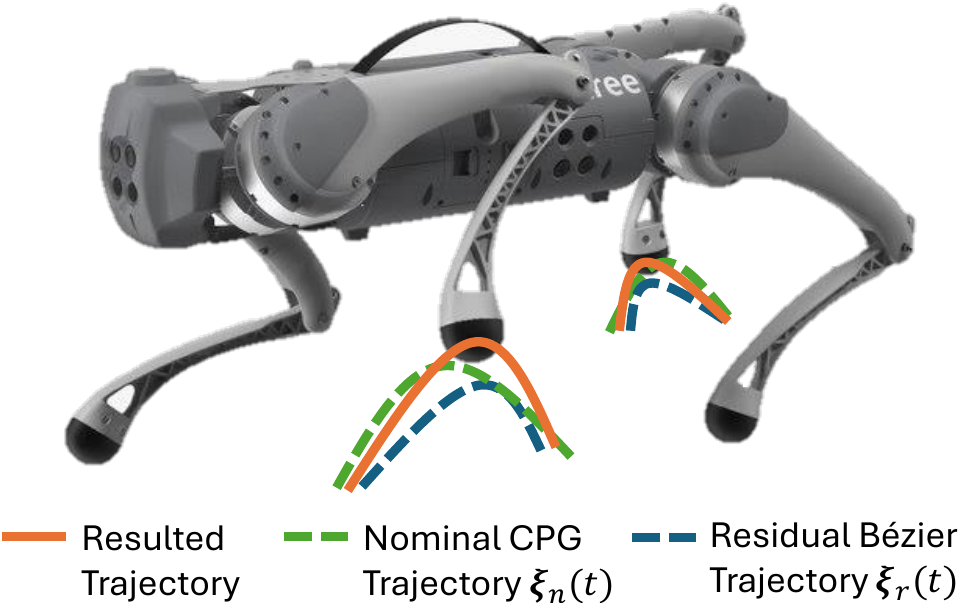}
\caption{In this work, the quadrupedal robot employs a trotting gait, characterized by two diagonal legs swinging simultaneously. For each swing leg, there is a desired end-effector trajectory, which is a summation of a nominal CPG trajectory $\bm{\xi}_n$ and a residual B\'ezier trajectory $\bm{\xi}_r$. By learning to change the control points of B\'ezier trajectories and the base movement, the policy could adjust the swing trajectories to perform various loco-manipulation tasks.}\label{fig:ee_traj}
\vspace{-0.3cm}
\end{figure}

We first provide a parameterized reference trajectory. 
Such a parameterized reference trajectory is the addition of two parts: (1) nominal trajectories for a trotting gait from a Central Pattern Generator (CPG), and (2) residual trajectories represented by B\'ezier trajectories that are added to the nominal trajectories, as illustrated in Fig.~\ref{fig:ee_traj}.
Given a trotting gait, there are two nominal and residual trajectories for the two swing legs, respectively. 
All the trajectories are defined in the base footprint frame of the robot. 
Having this design for the trajectories is the key to developing a versatile framework to enable the robot to learn to walk while adjusting the swing leg trajectories for manipulation tasks. 

\subsubsection{Nominal Trajectories from CPG}
The Central Pattern Generator (CPG) is a bio-inspired way to obtain swing foot trajectories, which is commonly used in quadrupedal locomotion control. 
CPG represents the trajectories of all four feet in 3D Cartesian space by high-order polynomials that are parameterized by desired base velocity $\dot{q}^d_{x,y}$ and turning yaw rate $\dot{q}^d_\psi$, with a given step height at the apex for swing legs. 
By using a periodic oscillator, the contact sequence of each leg can be obtained by choosing different oscillator parameters, which results in different rhythmic gaits, such as trotting, pacing, and etc. 
For more details of CPG, we refer readers to~\cite[Sec. III]{shao2021learning}.
For simplicity, we only use CPG to develop a trotting gait by fixing the parameters of the oscillator in this work. 
In this way, with a fixed gait period $T_g$, by varying the given command, the periodic trajectories $\bm{\xi}_n(t)$ can be well defined with respect to the time $t$. 
These serve as nominal trajectories for the operational space locomotion controller. 
We further denote $\dot{q}^d_{x,y,\psi}$ as CPG parameters which could be varied in this work.

\subsubsection{Residual Trajectories Represented by B\'ezier Curves}
B\'ezier curves are utilized to specify the change of the nominal trajectories from CPG, representing the residual trajectories $\bm{\xi}_r(t)$.
In this work, the B\'ezier curves are parameterized by three control points $\mathbf{P}_i$ ($i=0,1,2$) in 3D Cartesian space. 
To ensure the swing legs' trajectories start and end on the ground (with vertical displacement $z=0$), we set the vertical placement of the first and last control points to zero, leveraging the characteristic of Bézier curves to always pass through these points.
In this way, the B\'ezier parameters that can be chosen are $\mathbf{P}_{0,2}\in \mathbb{R}^2$ and $\mathbf{P}_1\in \mathbb{R}^3$.  
Then the change of position in the trajectory at time $t$ can be well defined by the B\'ezier function $\bm{\xi}_r(t)$. 
The time $t$ needs to be first normalized to be within $[0,1]$ w.r.t. the timespan of the B\'ezier trajectory (\textit{i.e.}, the robot swing period $T_{sw} = T_g / 2$).

\subsubsection{Control Objective} 
The objective of this locomotion controller at each time $t$ is to track the desired end-effector trajectories of the swinging legs, $\bm{\xi}_n(t) + \bm{\xi}_r(t)$, while tracking the commands for robot base, such as walking velocity and turning yaw rate (the CPG parameters $\dot{q}^d_{x,y,\psi}$). 

\subsection{Control Environment}
The control policy is trained via RL in a simulated environment to track target end-effector trajectories and robot base commands while trotting.

\subsubsection{Action}
The control policy's output actions specify desired motor positions $\mathbf{a}^c_k$ at each timestep $k$. The policy is queried at 50 Hz, and the target motor positions are then utilized in a joint-level PD controller running at 1 kHz to calculate motor torques.    

\subsubsection{Observation} 
At each timestep $k$, the control policy receives an observation from the environment and determines an appropriate action for that scenario. The observation $\mathbf{o}^c_k$ is the robot's proprioceptive feedback, which includes the robot's current angular position $q_{\phi,\theta,\psi}$, angular velocity $\dot{q}_{\phi,\theta,\psi}$, motor position $\mathbf{q}_m$, and motor velocity $\dot{\mathbf{q}}_m$. 
Alongside the current observation, we also keep track of a short 4-timestep history of the robot's past input (action) and output (observation), \textit{i.e.}, $(\mathbf{o}^c_{k:k-4},\mathbf{a}^c_{k-1:k-4})$, which are provided as input to the control policy. 
This short I/O history helps to facilitate state estimation and system identification. 
Additionally, we include a phase variable $\varphi=2\pi t/T_g$ for the control policy to indicate the time $t$ of one period of a trotting gait $T_g$. The phase variable is encoded using a smooth sinusoidal representation $(\sin{\varphi},\cos{\varphi})$.

\subsubsection{Goal}
In addition to the observations, the policy is also conditioned on a goal, which consists of two components. The first component is the desired end-effector trajectories to track, which are altered by B\'ezier parameters $\mathbf{P}_{0,1,2}$ for each swing leg. 
The second part includes the command for the robot's moving base, including desired sagittal velocity $\dot{q}^d_x$, lateral velocity $\dot{q}^d_y$, and turning yaw rate $\dot{q}^d_{\psi}$, which are also parameters for the CPG. 

\subsubsection{Reward}
At each timestep $k$, the robot executes an action $\mathbf{a}_k$ from the control policy. The environment then transitions to a new state and produces a reward $r^c_k$. 
The reward is a weighted composition of different terms.
The most dominating term (with the largest weight) is the end-effector tracking reward, which encourages the robot to track the desired trajectories of the four feet.
For swing legs, the reference trajectories are obtained in Sec.~\ref{subsec:ref_traj}. 
For stance legs, the reference trajectories become a single point to keep the foot unmoved. 
This reward is formulated as $r^c_\text{tracking} = \exp(-\sigma_e||\mathbf{x}_{e,t}-(\bm{\xi}_n(t)+\bm{\xi}_r(t))||_2)$ where $\mathbf{x}_{e,t}$ is the robot end-effector position $\mathbf{x}_e$ at time $t$ and $\sigma_e>0$ is a hyperparameter. 
Furthermore, penalties are also introduced for roll, pitch angular velocities and vertical linear velocity to stabilize the robot base and yaw position tracking. 

\subsubsection{Task Randomization}
To diversify training, we randomize the controller policy's tasks by varying the goal, by sampling B\'ezier parameters and CPG parameters uniformly from ranges in Table~\ref{tab:ctrl_point}.%
Diverse trajectories are generated by altering these parameters every 10 seconds. 
Training policies to perform a diverse range of tasks can also improve the robustness of the learned models~\cite{li2024reinforcement}.

\begin{table}
  \centering
  \begin{tabular}{cccc}
    \toprule
    \multicolumn{4}{c}{\textbf{B\'ezier Parameters Randomization Range}} \\
    \midrule

    Control points & $x$ Range [m]& $y$ Range [m]& $z$ Range [m]\\
    \midrule
    $\mathbf{P}_0$ & [-0.07, 0.03] & [-0.1, 0.1] & -\\
    $\mathbf{P}_1$ & [-0.035, 0.035] & [-0.1, 0.1] & [-0.05, 0.05]\\
    $\mathbf{P}_2$ & [-0.03, 0.07] & [-0.1, 0.1] & -\\

    \midrule
    \multicolumn{4}{c}{\textbf{CPG Parameters Randomization Range}} \\
    \midrule

    Command & Range & Unit & - \\
    \midrule
    $\dot{q}^d_x$ & [-1., 1.] & m/s & -\\
    $\dot{q}^d_y$ & [-0.3, 0.3] & m/s & -\\
    $\dot{q}^d_\psi$ & [-1., 1.] & rad/s & -\\
    \bottomrule
  \end{tabular}
  \caption{Range of B\'ezier parameters (control points) and CPG parameters used during training. }
  \label{tab:ctrl_point}
  \vspace{-5pt}
\end{table}

\subsubsection{Dynamics Randomization}
Since the training environment is built in simulation,  we also include dynamics randomization~\cite{peng2018sim} to facilitate transfer from simulation to the real world. The randomized dynamics parameters include joint PD gains, ground friction, base mass, and random perturbations, whose ranges are detailed in ~\cite[Table 3]{feng2023genloco}.

\subsection{Training and Deployment Details}
The episode is designed to last 4000 timesteps or 80 seconds, trained with GPU-accelerated Isaac Gym physics simulator. The policy uses a three-layer MLP with hidden sizes of $[128, 64, 32]$ and ELU activation, optimized by PPO~\cite{schulman2017proximal}. 
After convergence, the obtained locomotion controller can be zero-shot transferred to the robot hardware.

\section{Task-Specific Quadrupedal Manipulation Planning for Residual Trajectories}\label{sec:planning}
After obtaining the operational space locomotion controller that tracks arbitrary end-effector trajectories on the quadruped hardware, we can reuse this controller for different downsteam manipulation tasks.  
Given the diverse nature of the manipulation tasks, as described in Sec.~\ref{sec:framework}, we choose to develop task-specific high-level planning policies to specify the goal for the task-independent control policy.

\subsection{Planning Environment}
In this work, we use RL to develop the planning policy. 
As a note, HiLMa-Res allows us to leverage alternative approaches, like learning from demonstrations, for solving manipulation planning.

\subsubsection{Task-independent Action}
The manipulation planner determines the end-effector trajectories and robot base movement for the controller based on the given task. 
Specifically, the planner action $\mathbf{a}^p_k$ at each timestep includes B\'ezier parameters for swing legs' trajectories (control points $\mathbf{P}_{0,1,2}$) and CPG parameters (desired base velocity and turning rate $\dot{q}^d_{x,y,\psi}$). 
The selection of Bézier parameters (for residual trajectories) solely influences end-effector trajectories, while CPG parameters (for nominal trajectories) alter desired base movement.
This action space ensures compatibility with a task-independent locomotion controller across various tasks.

\subsubsection{Task-independent Observation}
The planner's observation consists of task-independent elements consistent across tasks, and task-specific elements unique to individual tasks. 
Task-independent observations include the robot base's measurable feedback, including its angular position and velocity, and the phase pair $(\sin{\varphi},\cos{\varphi})$ for synchronizing with the low-level locomotion controller's periodic gait.

\subsubsection{Additional Task-specific Observation} 
Each task-specific planner requires unique observations from the environment, either state or sensory feedback. 
HiLMa-Res allows us to flexibly manage both observation types.
\paragraph{State Representation}
In \emph{Dribble} task, the robot uses the relative position of the ball and the global heading of the robot to dribble the ball, conditioned on the given desired ball velocity (both speed and angle).
For \emph{Dribble}, where controlling the ball's velocity is crucial, we input a 4-timestep history of the robot and ball positions into the policy to infer the ball velocity without relying on noisy velocity estimates. 
In \emph{NavLoad} task, the robot is provided with robot pose and load position, conditioned on the navigation target, all in the global frame, to navigate the load to the target. 
We do not provide history as input in this position-based task. 

\paragraph{Vision Representation}
In the task of \emph{StepOStone}, obtaining an accurate state estimator and acquiring a precise height map that correctly labels the scattered stones could be difficult on a quadrupedal robot with onboard sensors. 
Therefore, the robot directly leverages its raw and downsampled depth vision from the ego view to learn to extract features related to the task.

\subsubsection{Reward}
The planning agent's reward $r^p_k$ at each timestep $k$ is task-specific within the hierarchical framework. 
Thanks to the HiLMa-Res, we are able to provide \emph{simple} task objectives without worrying the complexity of motion control, streamlining task learning and minimizing complex reward-tuning efforts present in end-to-end methods.

\paragraph{Dribble} The reward is to encourage minimizing the tracking error of the desired linear velocity and moving direction of the ball. We also include a proximity term to encourage the robot to stay close to the ball. 
Specifically, $r^p_\text{close} = \exp(-\sigma_{\text{ball}}d_{\text{ball}})$ where $d_\text{ball}$ is the distance between the robot and the ball and $\sigma_{\text{ball}}>0$ is a hyperparameter.  

\paragraph{StepOStone} The rewards are given for not stepping on stones or contacting with its calf. Specifically, $r^p_\text{contact} = 0$ if any leg is in contact, $1$ otherwise. Similarly, $r_\text{stepping} = 0$ if any leg steps on an object, $1$ otherwise. The weights are $0.7, 0.3$ respectively. 

\paragraph{NavLoad} We encourage the change of the distance between the load and the goal location. 
$r^p_\text{move\_load} = \Delta d_\text{load} = d_{k-1,\text{load}} - d_{k,\text{load}}$, where $d$ is the Euclidian distance between the goal and the load. 
We also include a proximity term formulated the same as the dribbling task, \textit{i.e.}, $r^p_\text{close} = \exp(-\sigma_{\text{load}}d_{\text{load}})$.

\subsection{Training for Different Tasks}
Using the HIMLA-Res, we explore different setups of the RL training, ranging from more capable but data-hungry algorithms that are mainly limited to simulation to data-efficient methods that can leverage real-world data.

\subsubsection{Tasks for State-based Policy}
Since the \emph{Dribble} planner incorporates only a short history of the ball's position, we use an MLP with three hidden layers of dimensions $[128, 64, 32]$ and ELU activation. Trained via PPO in Isaac Gym, we simulate a dragging force on the ball introduced in~\cite{ji2023dribblebot} to facilitate zero-shot transfer from simulation to the real world.

\subsubsection{Tasks for Vision-based Policy}
Since the \emph{StepOStone} policy directly uses depth vision, we firstly process the visual input via a CNN encoder with the first layer being convolution of size ([kernel, filter]) $[5, 32]$, second layer being Max Pooling of size ([kernel, stride]) $[2, 2]$, third layer being convolution of $[3, 64]$, ReLU activation, and a linear layer that maps the visual data into a latent space of hidden size of 64. Temporal features are later captured by a GRU with the same hidden size. It is then combined with state-based observation, and further processed by two MLP layers with hidden sizes of $[64, 32]$ and ELU activation. 
Unlike prior works that have to first train in state space and then transition to vision via teacher-student methods due to their low sample efficiency~\cite{kareer2023vinl,agarwal2023legged}, our efficient controller allows for direct end-to-end training with depth input in simulation by PPO in just five hours wall time. 
After training, the vision-based planner can be directly transferred to the real world. 

\subsubsection{Tasks that Requires Real-World Data}
In the \emph{NavLoad} task, we use a data-efficient DroQ~\cite{hiraoka2021dropout} algorithm to learn from real-world data. 
The actor network is an MLP whose first hidden layer has 128 ReLU units and the second has 64 Tanh units.
The critic network is a larger MLP with $[256, 128]$ hidden neurons, ReLU activation, and LayerNorm before the activation function. 
we leverage RLPD~\cite{ball2023efficient,smith2023learning} to fasten the real-world training.
First, we train the policy in simulation with DroQ (denoted as the base policy) and use this base policy to perform rollouts in the real world. 
We collect a replay buffer of 10,000 transitions. 
Then, we train a new policy from scratch using DroQ whose replay buffer is combined with both data collected by the base policy and the newly collected data from the real world, with a ratio of $70\%:30\%$. 
This could facilitate efficient learning in the real-world environment as described in~\cite{ball2023efficient}.

Till now, details of the development of the HiLMa-Res have been introduced. 

\section{Experiment Results}\label{sec:results}
In this section, we evaluate the HiLMa-Res policies for different loco-manipulation tasks extensively in both simulation and the real world. 
Due to the space limitation, we use \emph{NavLoad} task as an example for extensive benchmark, as it is more challenging, and use \emph{Dribble} and \emph{StepOStone} as extension examples in the real world.  

\subsection{Navigating a Load}
\figpushbox
\figuturn

We first benchmark the performance of different methods for the \emph{NavLoad} task in high-fidelity Gazebo simulation which provides a controlled environment. 
Then, we report real-world training results, followed by a discussion on the benefits of HIMLA-Res compared to the baseline methods. 

\subsubsection{Baseline}
We compare the HIMLA-Res method against three popular \emph{end-to-end} baselines on the \emph{NavLoad} task. The baselines include: 

\begin{itemize}
  \item \emph{Reward Shaping}: This baseline adopts a reward scheme from \cite{rudin2022learning} without reference motions and adds task reward directly to the original reward terms. 
  \item Adversarial Motion Prior (\emph{AMP}): This baseline is adopted from \cite{escontrela2022adversarial} and leverages adversarial imitation learning to learn from reference motions. 
  \item Motion Tracking (\emph{MT}): This baseline learns locomotion by tracking a given end-effector trajectory while improving the task reward~\cite{li2024reinforcement}. In implementation, we adopt the locomotion controller of HIMLA-Res without Bezier residuals and CPG commands, and add task-specific observation and reward for this specific task. 
\end{itemize}

These baselines reflect the state-of-the-art end-to-end locomotion controller, which has the potential to be extended to individual loco-manipulation tasks using task-specific policy.
\emph{We train each policy until it converges and achieves comparable success rates in training environments (Isaac Gym).} 

\figstepoverstone
\subsubsection{Simulation Benchmarking}
First, we benchmark the performance of HIMLA-Res against other methods in a high-fidelity \emph{Gazebo} simulation. Each method is tested for four different goals (1-meter in front, behind, left, or right of the load) with two attempts per goal, while the robot starts 1-meter behind the load at each reset. 
As shown in Table~\ref{tab:push-box}, HIMLA-Res outperformed all end-to-end comparison methods in success rates. Specifically, the \emph{Reward Shaping} method struggles to perform effective movement patterns during the sim-to-sim transfer, quickly losing balance and only kicking the load by chance. \emph{AMP} manages to move when the load is distant but faces difficulties in stabilizing the body (being confused about which motion the robot should exert) as it approached the load, likely due to mode collapsing in adversarial learning. The \emph{MT} baseline can move the loads to some extent but has trouble balancing motion imitation and task rewards, leading to unstable learning and evaluation performance. 
In contrast, HIMLA-Res, with its stable locomotion controller and focused task planner, completes the task without issues, highlighting the advantage of a hierarchical approach for complex tasks.

\begin{table}[t]
\centering
\begin{tabular}{l|c|c}
\toprule
\textbf{Method} & \textbf{Success Rate (\%)} & \textbf{Average Time (s)}  \\
\midrule

\multicolumn{3}{c}{Sim-to-Sim Transfer} \\

\midrule
\textbf{Reward Shaping}
    & 25 & 1.9 \\

\midrule

\textbf{AMP}
    & 0 & / \\

\midrule

\textbf{Motion Tracking}
    & 62.5 & 40.94 \\

\midrule

\textbf{Ours}      
    & \textbf{100} & \textbf{23.2}\\

\midrule
\multicolumn{3}{c}{Real World Experiments} \\

\midrule
\textbf{Ours (Base Policy)}      
    & 80 & 57.75\\

\midrule

\textbf{Ours (Trained in Real)}      
    & \textbf{100} & \textbf{47.6}\\

\bottomrule
\end{tabular}
\caption{Benchmark of the \emph{NavLoad} task in sim-to-sim transfer and real experiments. We find that all end-to-end methods cannot achieve a high success rate in simulation other than the one it was trained on, highlighting the difficulty of achieving both robustness and high task performance in end-to-end training. In comparison, our method achieves a 100\% success rate in simulation and can be transferred at an 80\% success rate zero-shot in real environments. Furthermore, we are able to achieve a 100\% success rate in the real environment after training with real data from scratch directly with RLPD algorithm. }
\label{tab:push-box}
\vspace{-5pt}
\end{table}

\begin{table}[t]
\centering
\begin{tabular}{l|c|c|c}
\toprule
\textbf{Trial \#} & \textbf{Stones Passed} & \textbf{Stones Fell}   & \textbf{\% Fell}\\

\midrule
\textbf{1}
    & 22 & 2 & 9.09\\

\midrule

\textbf{2}
    & 20 & 3 & 15\\

\midrule

\textbf{3}
    & 19 & 2 & 10.5 \\

\midrule

\textbf{4}
    & 19 & 3 & 15.8 \\

\midrule

\textbf{Average}
    & \textbf{20} & \textbf{2.5} & \textbf{12.5} \\

\bottomrule
\end{tabular}

\caption{Results of four trials on the \emph{StepOStone} task in real-world experiments. We report the number of stones the robot passed, and the number of stones which the robot steps on or hits. The robot avoids on average $87.5\%$ of the stones.}
\label{tab:steopovestone}
\vspace{-5pt}

\end{table}

\subsubsection{Real World Training and Evaluation}
We further test HiLMa-Res for \emph{NavLoad} in the real world. 
In the experiments, we utilize a localization system using Ultra-wideband (UWB) to measure the robot and load's positions and the robot's heading in the global framework. 
When we apply the simulation-trained base policy to real hardware, a notable drop in both success rates and efficiency is witnessed, due to the load's shape not being a perfect cube as in the simulation and the high noise in localization when the robot gets close to the load, leading to inaccurate pushes and frequent misses, as shown in Fig.~\ref{fig:pushbox}(a). 
To address these challenges, we utilize the RLPD algorithm to train with real-world data. 
The training from scratch with the RLPD algorithm takes just 9 minutes and 1,000 steps, leading to notable improvements in success rates and efficiency for real-world load navigation tasks. 
Compared to the base policy, which frequently misses the load and makes wide and inefficient turns, the policy trained with real-world data navigates more carefully. 
Shown in Fig.~\ref{fig:pushbox}(b), this strategy increases effective kicks and successful load interactions. 
On missing the load, it also adjusts with tighter turns for faster retries, showing better robustness to sensor noise. 
These result in a shorter path to reach the target. 
This efficiency underscores the benefits of a hierarchical approach, enabling efficient and reliable real-world task learning.

\subsection{Ball Dribbling}
In this \emph{Dribble} task, we use the robot onboard stereo fisheye cameras to detect the ball and calculate its relative position. 
With a proprioceptive state estimator, we can determine the robot's heading in the world frame for this short-term task. Our results as shown in Fig.~\ref{fig:uturn} demonstrate the dribbling planner's effectiveness, capable of executing a sharp U-Turn with the ball in spaces as narrow as 3.3 meters. The heading profile indicates a smooth approach to the target direction. 
Despite a significant ball detection error at 16 seconds, the planner is still robust and goes directly to the ball's position, with the help of its history observations that help to filter noisy measurements. 
The robot can dribble the ball at varying speeds, though the error in velocity tracking is large due to the gap in the physical properties of simulated and real soccer balls. Without a more accurate measurement system, we cannot learn the task directly in the real world due to noisy reward functions. 

\subsection{Step Over Stone}
In the \emph{StepOStone} task, we utilize a depth camera, downsampled to $43 \times 29$ to match our training setup, and apply a simulation-trained policy directly in real-world tests. 
These tests involve randomly placed wood blocks ranging from $3$cm to $5$cm wide and $5cm$ to $10cm$ tall on a padded surface.  
In Fig.~\ref{fig:stepostone}, on the center left snippets, the robot exhibits deliberate toe trajectories to bypass a tall wood block by adjusting the CPG trajectories (nominally 10cm high, risking collision) using B\'ezier residuals, enabling it to step over and land safely beyond the obstacle. 
On the center right snippets, the planner adeptly retracts the left front leg, preventing collision with another tall block, unlike the expected path of the nominal curve. 
Quantitatively, the robot can $87.5\%$ of the wood blocks on average across four trials, as recorded in Table~\ref{tab:steopovestone}. 
These results demonstrate the proposed framework's capability in mastering precise foot placements, notably in the \emph{StepOStone} task, and underscore the locomotion controller's accurate end-effector tracking.

\section{Conclusion and Future Work}
In conclusion, we present HIMLA-Res, a general hierarchical RL framework to tackle the loco-manipulation tasks. 
It combines a task-independent operational space locomotion controller that enables tracking control for both robot moving base and end-effector trajectories, and task-specific manipulation planners for downstream tasks. 
The HIMLA-Res is general and solves different loco-manipulation tasks.
Through these tasks, we show the flexibility of the HIMLA-Res framework in dealing with real-world training, agile loco-manipulation, and precise foot placements with the help of the policy hierarchy and the B\'ezier residual learning. 
Compared to end-to-end methods, HIMLA-Res refrains the planner from overcoming sim-to-real challenges and benefits it with significantly higher sample efficiency. 

Not limited to the presented tasks in this work, HiLMa-Res has the potential for other tasks, such as button pushing. 
We also note that the framework can also be extended to different quadrupedal gaits other than trotting by training with more CPG parameters such as different contact patterns or stepping heights. 
In the future, we think HiLMa-Res could also be useful for solving loco-manipulation tasks using humanoid robots.

\section*{Acknowledgements} 
This work was in part supported by The AI Institute and InnoHK of the Government of the Hong Kong Special Administrative Region via the Hong Kong Centre for Logistics Robotics. The authors thank the DAMODA Co., Ltd provides the UWB system.

{
\bibliographystyle{IEEEtran}
\bibliography{bib/bibliography}
}

\begin{acronym}
\acro{HP}{high-pass}
\acro{LP}{low-pass}
\end{acronym}

\end{document}